\pdfoutput=1

\documentclass[11pt]{article}

\usepackage[]{acl}

\usepackage{times}
\usepackage{multirow}
\usepackage{latexsym}
\usepackage{graphicx}

\usepackage[T1]{fontenc}

\usepackage[utf8]{inputenc}

\usepackage{microtype}
\usepackage{upgreek}
\usepackage{cleveref}
\usepackage{amsfonts}
\usepackage{algpseudocode}
\usepackage{algorithm}
\usepackage{float}
\newfloat{algorithm}{t}{lop}

\title{MicroBERT: Effective Training of Low-resource Monolingual BERTs through Parameter Reduction and Multitask Learning}

\author{Luke Gessler \quad Amir Zeldes\\
  Department of Linguistics \\
  Georgetown University \\
  \texttt{\{lg876,amir.zeldes\}@georgetown.edu} \\}

\begin{document}
\maketitle

\begin{abstract}
    Transformer language models (TLMs) are critical for most NLP tasks, but they are difficult to create for low-resource languages because of how much pretraining data they require.
    In this work, we investigate two techniques for training monolingual TLMs in a low-resource setting: greatly reducing TLM size, and complementing the masked language modeling objective with two linguistically rich supervised tasks (part-of-speech tagging and dependency parsing).
    Results from 7 diverse languages indicate that our model, MicroBERT, is able to produce marked improvements in downstream task evaluations relative to a typical monolingual TLM pretraining approach.
    Specifically, we find that monolingual MicroBERT models achieve gains of up to 18\% for parser LAS and 11\% for NER F1 compared to a multilingual baseline, mBERT, while having less than 1\% of its parameter count.
    We conclude reducing TLM parameter count and using labeled data for pretraining low-resource TLMs can yield large quality benefits and in some cases produce models that outperform multilingual approaches.
\end{abstract}

\def\CodeRepo{\url{https://github.com/lgessler/microbert}}
\def\microbert{$\upmu$\textsc{bert}}
\def\microberta{$\upmu$\textsc{bert-m}}
\def\microbertb{$\upmu$\textsc{bert-mx}}
\def\microbertc{$\upmu$\textsc{bert-mxp}}
\def\milliberta{$\upmu$\textsc{bert4-m}}
\def\millibertb{$\upmu$\textsc{bert4-mx}}
\def\millibertc{$\upmu$\textsc{bert4-mxp}}
\def\wordtovec{\textsc{word2vec}}
\def\mbert{\textsc{mbert}}
\def\mbertva{\textsc{mbert-va}}
\def\Thresher{%
\url{https://github.com/lgessler/wiki-thresher}
}

\section{Introduction}
Pretrained word embeddings are an essential ingredient for high performance on most NLP tasks.
Transformer language models (TLMs)\footnote{Following popular usage, we will informally refer to TLMs similar to the original BERT as ``BERTs'' throughout this work.} such as BERT/mBERT \citep{devlin_bert_2019} RoBERTa/XLM-R, \citep{liu_roberta_2019,conneau_unsupervised_2020}, and ELECTRA \citep{clark_electra_2020} provide state-of-the-art performance, but they expect at least tens of millions of tokens in training data. 
High-resource languages like English, Arabic, and Mandarin are able to meet this requirement, but most of the world's languages cannot.
Two major lines of work have arisen in order to address this gap: the first attempts to use multilingual transfer to pool different languages' data together to meet TLMs' data demands, and the second attempts to lower TLMs' data demands by changing their architectures and training regimens.

In this study, we take up work in the latter direction, asking specifically whether (1) vast reduction of model size and (2) incorporation of explicitly supervised, rather than self-supervised, tasks into model pretraining can produce better monolingual TLMs.
The former method is motivated by the intuition that normal-sized TLMs are so large as to be severely overparameterized for low-resource settings, and the latter method is motivated by an intuition that in the absence of large volumes of unlabeled text, signal from a supervised task with linguistic annotations is less likely be redundant to the model.
We find evidence that indicates both methods are helpful: our MicroBERT models produce monolingual embeddings that can outperform comparable multilingual approaches.
We summarize our contributions as follows:
\begin{enumerate}
\item We describe a method for training monolingual BERTs for low-resource settings, MicroBERT, characterized by a small parameter count and multitask pretraining which includes masked language modeling (MLM), part-of-speech (PoS) tagging and dependency syntax parsing.
\item Using evaluations on named-entity recognition (NER) and Universal Dependencies (UD) parsing across 7 diverse languages, we show that this approach is competitive with multilingual methods and often outperforms them for languages unseen by mBERT, even when the only pretraining task is MLM.
Our evaluation reveals a 7\% higher parser LAS and 6\% higher NER F1 on average for unseen languages, with gains up to 18.87\% and 11.81\% for parsing and NER. 
\item We release all MicroBERT models trained for this work at \CodeRepo{}.
\item We publicly release our code at \CodeRepo{} for reproducing our results and as a turnkey facility for training new MicroBERTs.
\end{enumerate}

\section{Previous Work}

At least since the development of pretrained static word embeddings \citep{mikolov_distributed_2013,mikolov_efficient_2013,pennington_glove_2014,bojanowski_enriching_2017}, pretrained word representations have been indispensable resources for NLP models, providing dense numerical representations of tokens' linguistic properties.
Pretrained contextualized embeddings \citep{mccann_learned_2018,peters_deep_2018,devlin_bert_2019} based on the Transformer architecture \citep{vaswani_attention_2017} have since overtaken them in popularity.
Throughout this period, high-resource languages have received the majority of attention, and although interest in low-resource settings has increased in the past few years, there remains a large gap (in terms of linguistic resources, pretrained models, etc.) between low- and high-resource languages \citep{joshi-etal-2020-state}.

\subsection{Multilingual Models} The publication of BERT \citep{devlin_bert_2019} also included a multilingual model, mBERT, trained on 104 languages.
mBERT and other massively multilingual models such as XLM-R \citep{conneau_unsupervised_2020} achieve impressive performance not just on those 104 languages but also in some zero-shot settings (cf., {\it inter alia}, \citealt{pires_how_2019,rogers_primer_2020-3}), despite the fact that models like mBERT do not have any explicit mechanism for inducing shared representations across languages.
However, large language models like XLM-R suffer from the fact that languages necessarily compete for parameters, meaning that barring fortuitous synergies each additional language should tendentially degrade the overall performance of the model for a fixed parameter count.
Moreover, languages with less training data tend to perform more poorly in LMs like XLM-R \citep{wu_are_2020}.

While the majority of multilingual models seek to include many languages, with a large proportion of them being high-resource, there are some low-resource approaches to training multilingual models from scratch where there may not even be any high-resource languages. 
For example, \citet{ogueji_small_2021} train an mBERT on data totaling less than 1GB ($\approx$100M tokens) from 11 African languages, and find that their model often outperforms comparable massively multilingual models.

\subsection{Adapting Multilingual Models}
One response to the difficulties posed by massively multilingual models has been to leave aside the goal of fitting ever more languages into a single model, and to investigate whether it would be more fruitful to {\it adapt} pretrained massively multilingual models for a given target language.
Enriching the TLM's vocabulary with additional tokens (e.g.~wordpieces for BERT-style models) has been shown to be helpful because of how it improves tokenization and reduces the rate of out-of-vocabulary tokens \citep{wang_extending_2020,artetxe-etal-2020-cross,chau_parsing_2020,ebrahimi_how_2021}.
Transliteration has also been shown to be beneficial when there are related languages that would not have been able to benefit from transfer in the form of shared representations otherwise, e.g.~between Turkish (Latin script) and related Uyghur (Arabic script) \citep{muller_when_2021,chau_specializing_2021}.
Using adapter modules \citep{houlsby_parameter-efficient_2019} has also proven effective \citep{pfeiffer_adapterhub_2020}. 
All these approaches are typically combined with {\it continued pretraining}, where MLM and other pretraining tasks are used to update model weights, and some formulations of continued pretraining are multitask \citep[][{\it inter alia}]{pfeiffer_mad-x_2020,chau_specializing_2021}.

\subsection{Monolingual Models}
Whereas multilingual approaches have tried to address low-resource settings with transfer from high-resource languages, other approaches have investigated the question of how much data is needed for a given level of quality in a BERT-like model, and the question of what alternative training regimens might help reduce this data requirement.

Several studies have examined notable thresholds on dataset size.
\citet{martin_camembert_2020} find in a series of experiments that for French, at least 4GB of text is needed for near-SOTA performance, and \citet{micheli_importance_2020} show further that at least 100MB of text is needed (again for French) for ``well-performing'' models on some tasks. 
\citep{micallef_pre-training_nodate} perform similar experiments for a monolingual Maltese BERT, finding that even when trained with only 46M tokens, the monolingual BERT, BERTu, was able to achieve results competitive with an mBERT model adapted with the vocabulary augmentation methods of \citet{chau_parsing_2020}.
\citep{warstadt_learning_2020} train English RoBERTa models on datasets ranging from 1M to 1B tokens and find that while models acquire linguistic features readily on small datasets, they require more data to fully exploit these features in generalization on unseen data.

To our knowledge, there has been little work on examining whether significantly reducing model size could help in the low-resource monolingual setting. 
As a baseline, \citet{chau_specializing_2021} and \citet{muller_when_2021} train monolingual BERTs with 6 instead of 12 layers for low-resource languages, but this does not even halve the model's parameter count.
The only exception we were able to find is work from \citet{turc_well-read_2019}, where very small models (as low as 4.4M parameters to BERT base's 110M) are pretrained directly prior to training via distillation, but the condition where the small model is only pretrained and not trained via distillation is not evaluated in their work.

\subsection{Non-TLM Models}
Finally, it is worth noting that while BERT-like TLMs are the clear winner overall for high-resource languages in most tasks, in low-resource settings, other embedding models may be superior.
\citet{arora_contextual_2020} and \citet{ortiz_suarez_monolingual_2020} show that ELMo \citep{peters_deep_2018}, static \citep{pennington_glove_2014}, and even random embeddings are often not too far behind BERT-like TLMs on some tasks even for high-resource languages.
\citet{riabi_can_2021} show that a character-based language model is competitive with mBERT for one low-resource language, NArabizi.

\section{Motivation}
As we have seen, monolingual BERTs trained with standard methods tend to perform poorly when less than 20-40M tokens are available during training, and there is evidence that they do not learn to fully generalize some linguistic patterns without a large ($\approx$1B tokens, \citealt{warstadt_learning_2020}) amount of training data.
However, most popular methods for pretraining BERTs are self-supervised, using only unlabeled text.
This has turned out well for high-resource languages, where unlabeled text is available in far greater quantities than labeled text, to the point where incorporating labeled text into pretraining does not always provide large gains.

However, even in very low-resource settings, it is common for sources of linguistic signal beyond unlabeled text to be available, such as treebanks, interlinearized text, and dictionaries.
It is natural to ask whether using them as data for auxiliary supervised tasks during model pretraining could help monolingual models overcome a lack of unlabeled data, and perhaps even interact synergistically with the main pretraining task, such as MLM.
It is known, for example, that BERTs learn to represent words' parts of speech \citep{rogers_primer_2020-3}, and it seems possible that providing direct supervision for predicting parts of speech may help a model acquire good PoS representations with less data.
This leads us to our first hypothesis \textbf{H1}, that monolingual models should benefit from multitask pretraining with auxiliary tasks incorporating labeled data.

Previous results also lead us to our second hypothesis \textbf{H2}, that in low-resource settings, monolingual BERTs are typically severely overparameterized.
Most BERTs are overparameterized in the sense that they can have modules removed, disabled, or compressed while showing minimal regressions (or sometimes even improvements) \citep{rogers_primer_2020-3}, but in H2 we mean further that there are so many parameters that the model cannot be effectively learned given the amount of data. 
As noted in \S2, there appears to be a gap in the literature on whether pretraining a vastly scaled down BERT model could help monolingual BERTs perform better in low-resource settings, and we take up the question in this work.

\section{Approach}
We propose an architecture and training regime for monolingual BERTs which we call MicroBERT. 
We keep the basic architecture of BERT, but we reduce encoder layer count to 3, hidden representation size to 100, and number of attention heads to 5.
(Compare this to BERT base's 12, 768, and 12, respectively.)
Excluding prediction heads, this reduces parameter count from 108M\footnote{Obtained from \texttt{\small bert-base-cased} using the \texttt{BertModel} implementation in HuggingFace's \texttt{transformers} library.} to 1.29M, or just 1.19\% of a normal BERT model's size.
After the encoder stack, one dedicated head is used for each task, where each head is provided with the last encoder layer's hidden states.

For training, assume a task set $\mathcal{T} = t_1, \ldots, t_{|\mathcal{T}|}$, corresponding datasets $\mathcal{D} = d_1, \ldots, d_{|\mathcal{T}|}$, and a set of weights for each task $\lambda_1, \ldots, \lambda_{|\mathcal{T}|}$, s.t.~$\sum_i\lambda_i=1$.
To prepare the sequences of batches $\mathcal{B} = b_1, \ldots, b_{|\mathcal{B}|}$ for a given epoch, construct each batch $b_i$ using only instances from exactly one dataset $d_t$, and sample batches so that each dataset $d_t$ is represented at least $\lfloor\lambda_t|\mathcal{B}|\rfloor$ times in $\mathcal{B}$.
Each batch is sent not only to its dataset's corresponding prediction head, but also to any other prediction heads which are compatible with it.
For example, a batch containing dependency syntax labels would be sent to the parsing prediction head, and it would also be sent to the MLM head, since the MLM head only requires unlabeled text.%
\footnote{Actually, matters are a bit more complicated than this. 
The MLM head requires representations that included a \texttt{[MASK]} token from the start, whereas other heads require representations from unmasked sequences.
For multitask batches, therefore, the batch must be fed through the encoder stack twice: once with masking, and once without masking.}
If a dataset is exhausted in the course of this procedure, new instances are sampled anew from the beginning of the dataset.
This is a simple means for addressing the fact that some datasets will be much larger than others, which without intervention could have led to one task's parameter updates drowning out others.

We consider three tasks in this work.
The first is MLM implemented as whole-word, dynamic masking, as in RoBERTa \citep{liu_roberta_2019}.
The second is PoS tagging, for which our prediction head is a simple linear projection.
The third is dependency parsing, for which we use a modified form of the biaffine dependency parser of \citet{dozat_biaffine_2017} which has had the encoder LSTM stack removed.
Cross-entropy loss is used for all tasks and summed together: each head produces an associated loss $\ell_i$, which is summed into a single loss $\ell$ which is used to begin backpropagation.
We note that it would be straightforward to add other tasks, though we choose PoS tagging and parsing for this work since PoS tagged and dependency parsed datasets are relatively common for low-resource languages.
This multitask setup is not novel---in fact, \citet{chau_specializing_2021} use the the same three tasks for a similar purpose, though instead of pretraining a BERT from scratch, they use the multitask setup to perform adaptive finetuning on a pretrained multilingual model, and find a negative result.

\section{Experimental Methods}

To evaluate our approach, we train MicroBERT models on several languages and compare them to a variety of baselines. 
All our experiments are implemented using AllenNLP \citep{gardner-etal-2018-allennlp}, Transformers \citep{wolf_huggingfaces_2020}, and PyTorch \citep{NEURIPS2019_9015}.
All code and models are available at \CodeRepo.

\subsection{Data}
We prepare datasets for seven diverse languages: Wolof, Uyghur, Ancient Greek, Maltese, Coptic, Indonesian, and Tamil. 
These languages were selected according to several criteria.
First, two hard requirements were that they needed to have a Universal Dependencies \citep{nivre_universal_2016} treebank with a train, dev, and test split; and that they needed to have a ``large-enough'' source of unlabeled text totaling between 500,000 and 10,000,000 tokens.
Second, languages were prioritized based on phylogenetic diversity: six unrelated language families are represented (Niger--Congo, Turkic, Indo-European, Afro-Asiatic, Austronesian, Dravidian), and languages vary widely in syntax (for example, Uyghur is morphologically rich, while Coptic is morphologically poor).
Third, we sample languages along the spectrum of data quality---for example, some have very high quality tokenization, while others have noisier tokenization.

\begin{table}[]
    \centering
    \small
    \begin{tabular}{l|rrr}
Language   & Unlabeled & UD & NER\\\hline
Wolof      &    517,237 &    9,581 &   10,800 \\
Coptic     &    970,642 &   48,632 &       -- \\
Tamil      &  1,429,735 &   40,236 &  186,423 \\
Indonesian &  1,439,772 &  122,021 &  800,063 \\
Maltese    &  2,113,223 &   44,162 &   15,850 \\
Uyghur     &  2,401,445 &   44,258 &   17,095 \\
Anc. Greek &  9,058,227 &  213,999 &       -- \\
    \end{tabular}
    \caption{Token count for each dataset by language, sorted in order of increasing unlabeled token count. Recall that unlabeled data for Indonesian and Tamil was downsampled, and all other sources of unlabeled data were used in full.}
    \label{tab:token_stats}
\end{table}

For each language, we obtain a UD treebank, a larger unlabeled corpus, and for all languages except Ancient Greek and Coptic, an NER dataset from WikiAnn \citep{pan-etal-2017-cross}. 
Unlabeled data for each language was taken from Wikipedia, except for Ancient Greek and Coptic, whose unlabeled corpora were taken from open access digital humanities projects.
Note that the unlabeled corpora for Indonesian and Tamil were downsampled by randomly choosing Wikipedia articles until a quota of around 1.5M tokens was met.
A summary of corpus statistics is given in \Cref{tab:token_stats}, and a full description of the languages' datasets and their preparation is given in \Cref{sec:datasets}.
Note that Uyghur is written in Arabic script; Wolof, Indonesian, and Maltese are written in Latin script; and Tamil, Coptic, and Ancient Greek are written in their own scripts.

\subsection{Conditions}
We compare four baselines, as well as six variants of the MicroBERT approach.

\begin{itemize}
\item \textbf{\wordtovec{}}: a 100-dimensional static word embedding baseline, motivated by observations that static word embeddings can perform well in low resource settings (cf. \S2).
\item \textbf{\mbert{}}: the \textit{bert-base-multilingual-cased} pretrained model. Note that only two of our seven languages (Indonesian and Tamil) have been seen by \mbert{}.
\item \textbf{\mbertva{}}: the \textit{bert-base-multilingual-cased} pretrained model adapted in the \textit{vocabulary augmentation} method of \citet{chau_parsing_2020}, where 99 wordpieces are added to the vocabulary and the model is pretrained further.
\item \textbf{\microberta{}, \microbertb{}, \microbertc{}}: our MicroBERT models with MLM; MLM and XPOS\footnote{In Universal Dependencies parlance, an XPOS tag is a part of speech tag from a language-specific tag inventory, as opposed to a UPOS, which is drawn from a universal tag inventory.} tagging; and MLM, XPOS tagging, and UD parsing used in pretraining. \microbertb{} performs tasks at an 8:1 ratio, and \microbertc{} performs tasks at an 8:1:1 ratio.
\item \textbf{\milliberta{}, \millibertb{}, \millibertc{}}: like the corresponding MicroBERT models, but approximately 4 times larger, having 200 instead of 100 hidden units; 8 instead of 5 attention heads; and 6 instead of 3 layers.

\end{itemize}

Our \microbert{} models are all trained for 200 epochs with a batch size of 32 and 8,000 batches per epoch, and we save the model that achieves best MLM performance on the validation split of the unlabeled dataset.
This results in our models being trained on only 20\% of the batches that BERT was, though we hypothesize that due to our smaller model and dataset sizes, this may not be an issue. 
A full description of our methods is given in \Cref{sec:conditions}.

\subsection{Evaluation}
To evaluate our pretrained models, we perform NER on the WikiAnn datasets and dependency parsing on the UD datasets for each language--model pair, following previous work \citep[][{\it inter alia}]{chau_parsing_2020,muller_when_2021}.
We choose these tasks because they are common in the literature of TLM evaluation, because datasets are common even in low-resource languages for them, and because they both assess somewhat complementary linguistic information: informally, parsing requires grammatical knowledge, and NER requires semantic and world knowledge.
Combined, they ought to give a holistic view of a model's abilities.

We use common hyperparameter settings to train the evaluation models which allow for fine-tuning of the BERT model at a reduced learning rate.
A standard \citet{dozat_biaffine_2017} parser is used for the parsing evaluation, and a linear chain CRF with stacked LSTM encoders is used for the NER evaluation.
Our metrics for these tasks are LAS and span-based F1 score
respectively. 
Gold tokenization is used in both evaluations. 
No auxiliary input signals (e.g.~PoS embeddings, morphological feature embeddings, static embeddings) are used.
We forgo auxiliary inputs even though they would likely improve our scores, and even though it means no longer being able to compare our performance directly to numbers reported in some other works, since we believe providing the model's representations as the sole input provides the clearest picture of its quality.%
\footnote{This is motivated by our experience in preliminary experiments of using a parser with these auxiliary inputs, with the result that differences between our models were no larger than 3\% since the auxiliary inputs were contributing so much to the model's performance, obscuring the content of the model representations.
We also notice a similarly small difference between comparable models in other works where auxiliary inputs were used in a parsing evaluation.}
Full descriptions of the evaluation models is available in \Cref{sec:evaluation}.

\def\tbd{\textsc{tbd}}
\section{Results}
\begin{table*}[ht]
    \centering
    \tiny
    \begin{tabular}{l|rrrrrrr|r}
                       & Wolof & Coptic & Maltese & Uyghur & An. Gk. & Tamil & Indon. & Avg. \\\hline\hline
        \wordtovec     & 72.35 & 85.69 & 73.41 & 54.27 & 73.30 & 50.91 & 74.10 & 69.15 \\
        \mbert         & 76.40 & 14.43 & 78.18 & 46.30 & 72.30 & \textbf{66.73} & {\bf 78.63} & 61.85 \\
        \mbertva       & 72.94 & 82.11 & 72.69 & 42.97 & 65.89 & 54.92 & 75.67 & 66.74 \\
        \citet{chau_specializing_2021} & 60.12 & & 65.92 & 60.34 & & & \\         
        \microberta    & 75.69 & 86.45 & 74.33 & 61.26 & 78.95 & 59.75 & 74.66 & 73.01 \\
        \microbertb    & \textbf{77.83} & \textbf{88.25} & 78.90 & \textbf{65.17} & \textbf{80.55} & 61.00 & 74.69 & \textbf{75.20} \\
        \microbertc    & 73.30 & 86.35 & 75.11 & 59.98 & 79.08 & 58.05 & 73.28 & 72.16 \\
        \milliberta    & 74.42 & 82.72 & \textbf{79.25} & 57.79 & 79.59 & 61.09 & 74.32 & 72.74 \\
        \millibertb    & 73.99 & 82.52 & 78.61 & 57.14 & 79.09 & 60.82 & 74.21 & 72.34 \\
        \millibertc    & 74.30 & 82.73 & 78.99 & 57.01 & 79.56 & 60.92 & 74.34 & 72.55 \\\hline
        \microbertb{} -- \mbertva & 4.89 & 6.14 & 6.21 & 22.20 & 14.66 & 6.08 & -0.97 & 8.46 \\
    \end{tabular}
    \caption{Labeled attachment score (LAS) by language and model combination for UD parsing evaluation. The final row shows the difference in score between \microbertb{} and \mbertva{}. Results from \citet{chau_specializing_2021}'s half-sized monolingual BERT are included for comparison.}
    \label{tab:parser_results}
\end{table*}

\begin{table*}[ht]
    \centering
    \tiny
    \begin{tabular}{l|rrrrr|r}
                       & Wolof & Maltese & Uyghur & Tamil & Indon. & Avg. \\\hline\hline
        \wordtovec     & 86.89 & 82.67   & 86.37  &  82.71 & 94.28 & 86.58 \\
        \mbert         & 83.79 & 73.71   & 78.40  &  70.47 & 91.04 & 79.48 \\
        \mbertva       & 79.37 & 78.11   & 77.03  &  69.38 & 91.05 & 78.99\\
        \microberta    & \textbf{83.92}  & 75.89  & 81.36  &  \textbf{82.28} & 92.25 & 83.14 \\
        \microbertb    & 81.12 & 84.80   & \textbf{85.45}  &  81.61 & 92.43 & 85.08 \\
        \microbertc    & 82.21 & \textbf{88.79}   & 82.52  &  82.00 & 92.27 & \textbf{85.56} \\
        \milliberta    & 78.69 & 78.22   & 80.28  &  80.57 & \textbf{93.05} & 82.16 \\
        \millibertb    & 80.95 & 80.00   & 79.36  &  80.12 & 92.55 & 82.60 \\
        \millibertc    & 79.02 & 79.31   & 81.59  &  80.11 & 93.01 & 82.61 \\\hline
        \microbertb{} -- \mbert{} & -2.67 & 11.09 &  7.05  &  11.14 &  1.39 & 5.60 \\
    \end{tabular}
    \caption{Span-based F1 score by language and model combination for NER evaluation. The final row shows the difference in score between \microbertb{} and \mbert{}. Boldface indicating top performance for a language does not consider \wordtovec.}
    \label{tab:ner_results}
\end{table*}

Results for the parser evaluation are given in \Cref{tab:parser_results}, and results for the NER evaluation are given in \Cref{tab:ner_results}.
For both tables, we also include additional rows comparing important model pairs.

It is possible to directly compare our parsing evaluation results with those of \citet[][Table 2]{chau_specializing_2021}, whose evaluation methodology we closely follow for parsing. 
For our three overlapping languages---Maltese, Uyghur, and Wolof---we find that LAS for mBERT is similar, which establishes that evaluation conditions are comparable.
We include their half-size BERT model's numbers in \Cref{tab:parser_results} for comparison, which were obtained by training a \texttt{bert-base}-sized BERT from scratch on the target language with 6 instead of 12 layers.

\paragraph{Non-DNN Baseline} 
First, corroborating prior work, we can see that static word embeddings are competitive for many languages, often outperforming the multilingual models in both tasks, and often performing best overall for NER.

\paragraph{Multilingual Baselines} 
Note the generally poor performance of \mbertva{}, which we had hoped would be a baseline stronger than \mbert{}, but often underperforms relative to \mbert{}.%
\footnote{An exception to this is parsing for Coptic, where \mbert{}'s lack of wordpieces for Coptic script causes a high out-of-vocabulary rate, giving \mbertva{} an obvious advantage.}
Poor performance persisted even after carefully ruling out implementation errors. 
We speculate that this was likely caused by the general instability inherent in fine-tuning large TLMs \citep{rogers_primer_2020-3} and the authors of \citet{chau_parsing_2020} point out in correspondence that the especially small batch size we were forced to adopt for \mbertva{} due to GPU memory constraints, 2, could have led to the degradation.
(It is a common finding that a large batch size is necessary to prevent noisy gradients for TLMs---cf.~\citealt[][inter alia]{clark_electra_2020,devlin_bert_2019}.)
We therefore view our experiments as inconclusive on whether \mbertva{} can yield strong results for these languages.

\paragraph{Monolingual Model Size}
We can see that for parsing and NER, the \microbert{}4 model performs worse in almost all cases than the equivalent \microbert{} model.
The degradation is -0.27\% on average for \textsc{-m} variants, and -2.86\% on average for \textsc{-mx} variants. 
The one language for which the \microbert{}4 model performs much better on parsing is Maltese, where the \milliberta{} model performs 5\% better than the \microberta{} model, indicating that in this experimental condition greater model size may help, though note that the Chau and Smith's half-BERT does much worse than \milliberta{} showing a 13\% lower score compared to \microberta{} and reversing the trend.
On our two other languages in common with Chau and Smith, we see an 18\% (Wolof) and 5\% (Uyghur) degradation relative to \microbertb{}.
For NER, we similarly observe that the \microbert{}4 variants have worse average performance than \microbert{} variants. 
We take this all to be strong evidence for H2, that monolingual BERTs trained at common sizes are severely overparameterized in low-resource settings, to the point that large performance degradations are observed. 

\paragraph{Parsing}
Considering the five languages unseen by mBERT (all except Tamil and Indonesian), we see in the parsing results that in every case the best monolingual model, usually \microbertb{}, is able to outperform the best multilingual model.
In some cases the difference is very large, such as in Uyghur parsing where there is an absolute gain in 18.87\% LAS, and in others it is within the range of chance, such as in Maltese parsing.
For the languages mBERT has seen, Tamil and Indonesian, \mbert{} outperforms the \microbert{} by several points, though we find it remarkable that \microbert{} is able to still provide a competitive score despite being trained on very small subsets of Tamil and Indonesian Wikipedia (150K and 600K articles, respectively), which mBERT had full access to.
\microbertb{} performs best of all the models, achieving a score 8.5\% higher than that of \mbertva{} on average.

\paragraph{NER} 
Turning now to NER results, we see that in three cases, our \microbert{} models are able to clearly outperform other models, including Tamil, which \mbert{} has seen.
In the other two cases, Indonesian and Wolof, \microbert{} models technically keep a lead but with margins thin enough to be noise.
For all languages except Maltese however, \wordtovec{} is able to meet or beat top performance from TLMs. 
Taken together with the parsing results, where \wordtovec{} underperforms, and with the strengths and weaknesses of contextualized and static embeddings in mind, we hypothesize that NER on the WikiAnn dataset may require rote capacities, such as name memorization, instead of sophisticated linguistic knowledge, especially on an automatically-constructed dataset like WikiAnn.

\begin{figure}
    \centering
    \includegraphics[width=\linewidth]{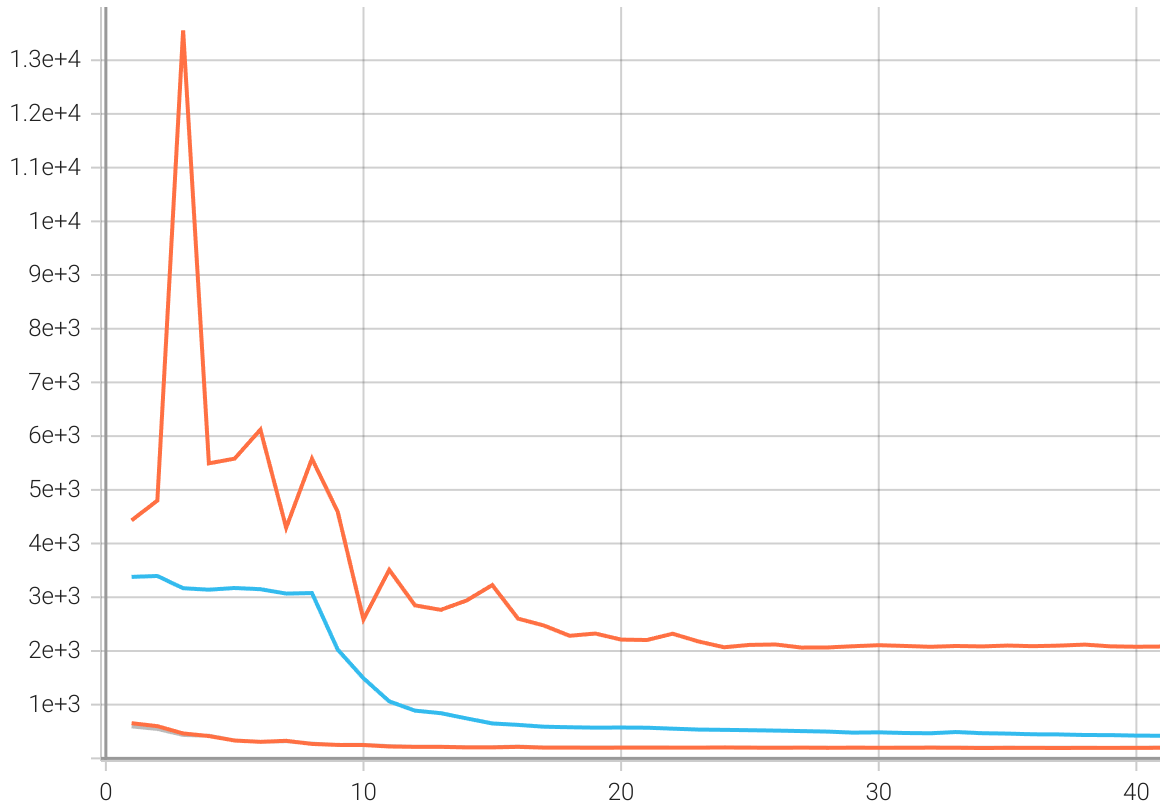}
    \caption{MLM perplexity vs. epoch for the validation split of the Uyghur dataset.
    The top line is \microberta{}, the middle line is \microbertc{}, and the lowest line is \microbertb{}.}
    \label{fig:uyghur_validation_perplexity}
\end{figure}

\paragraph{Validation MLM Perplexity} 
In order to better understand the effects of our auxiliary tagging and parsing tasks, we examine the validation MLM perplexity of our models during pretraining.
An example of these curves is given in \Cref{fig:uyghur_validation_perplexity}.
We first observe that for all languages, validation MLM perplexity is lower at all times for the multitask models compared to the perplexity curve for the MLM only model.
Moreover, validation MLM perplexity converges more quickly on its asymptotic value for multitask models.
For \microbertb{} in particular, validation MLM perplexity usually comes very close to its final value even within the first 10 epochs of pretraining.
Validation MLM perplexity is only one incomplete measure of model quality, and indeed it is not entirely predictive of downstream performance since \microberta{} sometimes outperforms \microbertb{} and \microbertc{}. 
But we take these results as evidence that our auxiliary tasks are helping our models learn more quickly. 
Moreover, while proving this would require additional work, it seems possible from the shapes of the validation curves that for the smallest datasets, multi-task learning (MTL) might be helping models learn more than they could have through MLM alone.

Within validation MLM perplexity, we also see that each language follows one of two patterns: either the perplexity curves for \microbertb{} and \microbertc{} are nearly identical, or the perplexity curve for \microbertc{} remains a bit higher than for \microbertb{}.\footnote{The former pattern holds for Wolof, Maltese, Greek, Indonesian, and Tamil, and the latter pattern holds for Uyghur and Coptic.}
With the intuition that more auxiliary tasks ought to make MLM easier, we had hypothesized that if anything the curve for \microbertb{} would have been higher than for \microbertc{}, but instead the reverse sometimes turned out to be true.
We hypothesize that the difference in task proportions between \microbertb{} and \microbertc{} might have been partially responsible for this: in the former, 1 in 9 batches are for auxiliary tasks, and in the latter, 2 in 10 batches are for auxiliary tasks.
If this is true, then finding the right proportion of primary and auxiliary tasks during pretraining would be critical for the multitask pretraining approach.

\section{Discussion}

\paragraph{Main Findings}
We take our most important result to be our demonstration that it is possible to train a monolingual BERT from scratch that can compete with and even outperform multilingual models by up to 18\% LAS and 11\% NER F1 using as little as 500,000 tokens and a UD treebank of 44,000 tokens and less than 1\% of the parameters.

\paragraph{Multilingual Baselines}
We chose to use mBERT as a baseline because it is widely used and well studied.
Moreover, given the the architectural homogeneity of mBERT and other leading multilingual LLMs, we additionally believe mBERT is strong enough to be representative of the state of multilingual LLMs for this work.
While \mbertva{} appeared to severely underperform in some cases, we observe that it was still a strong baseline for both tasks (on average 5\% better than \mbert{} for parsing, and 0.5\% worse).
In sum, while slightly stronger multilingual baselines may exist, we believe the ones in this work were still strong enough to show the MicroBERT approach holds promise, given that MicroBERTs were able to perform better than multilingual LLMs by several percentage points on average in both tasks.

\paragraph{Hypotheses} 
We find strong support in these results for H1, that monolingual TLMs often, though not always, benefit from multitask learning on labeled data in low-resource settings.
We additionally find strong support for H2, that when data is severely limited, typical BERT configurations are harmfully overparameterized.


\paragraph{Future Work}
There remain some unanswered questions in this work.
The addition of the third parsing task proved harmful to performance in most cases, and it is unclear why.
Parsing and XPOS tagging involve much of the same linguistic phenomena, and it seems possible that replacing one of them with a more semantic auxiliary task might have led to better results.
Another possibility is that having loss computed for auxiliary tasks only on {\it some} batches may lead to jerky or suboptimal paths along the loss gradient, a problem which could be mitigated by having batches where only some sequences are suitable for use in auxiliary tasks.


It is natural to ask whether any of the elements of our approach here could find use in multilingual settings.
Reducing the size of multilingual models may not be a promising direction due to the curse of multilinguality \citep{conneau_unsupervised_2020}. 
\citet{ogueji_small_2021} show further that even for low-resource multilingual models, size still seems to be important.
As for multitask learning, \citet{chau_specializing_2021} find a negative result for using MTL in multilingual model adaptation, though given the complex nature of MTL, many possible approaches remain untried.

Most languages in the world lack PoS tagged and parsed datasets, and if the MicroBERT approach is to be extended to very low-resource languages, it is likely that other auxiliary tasks would be needed.
We leave this direction to future work, though we speculate that there are plenty of alternatives that may work.
Parallel corpora, often in the form of a Bible translation, are readily available for over a thousand of the world's languages.
High-quality rule-based morphological parsers are sometimes available for very low-resource languages, and their outputs could be used like PoS tags.
Interlinearized texts and dictionaries are also common products of language documentation which are rich in linguistic information.
All of these resources could be adapted for use in an auxiliary task.


\section{Conclusion}

We have shown that it is possible to train monolingual TLMs that are competitive with multilingual models using as little as 500K tokens and a 40K token treebank with greatly reduced model size and multitask learning on PoS tagging and dependency syntax parsing.
While multilingual models did have some advantages over our approach, we observe that our MicroBERT approach has unique strengths for work on low-resource TLMs, including its lack of reliance on successful cross-lingual transfer and radically reduced computational demands for pretraining and downstream use.

We take this result to call into question whether multilingual representation learning can scale down effectively to truly ``low-resource'' languages that have less than a few million tokens in training data.
Sometimes languages like these can be well served by transfer from related languages, even if all languages are low-resource \citep{ogueji_small_2021}, but not all languages may be so lucky: language isolates by definition lack related languages, and small language families are likely less able to benefit from transfer, since transfer tends to be enabled by phylogenetic \citep{nguyen-chiang-2017-transfer} or areal \citep{goyal-etal-2020-contact} relatedness between languages.
While multilingual methods hold much promise, it is important to examine other approaches to low-resource representation learning which, if not strictly better, may at least be complementary.


\bibliography{anthology,custom}

\begin{thebibliography}{46}
\expandafter\ifx\csname natexlab\endcsname\relax\def\natexlab#1{#1}\fi

\bibitem[{Arora et~al.(2020)Arora, May, Zhang, and Ré}]{arora_contextual_2020}
Simran Arora, Avner May, Jian Zhang, and Christopher Ré. 2020.
\newblock \href {https://doi.org/10.18653/v1/2020.acl-main.236} {Contextual
  {Embeddings}: {When} {Are} {They} {Worth} {It}?}
\newblock In \emph{Proceedings of the 58th {Annual} {Meeting} of the
  {Association} for {Computational} {Linguistics}}, pages 2650--2663, Online.
  Association for Computational Linguistics.

\bibitem[{Artetxe et~al.(2020)Artetxe, Ruder, and
  Yogatama}]{artetxe-etal-2020-cross}
Mikel Artetxe, Sebastian Ruder, and Dani Yogatama. 2020.
\newblock \href {https://doi.org/10.18653/v1/2020.acl-main.421} {On the
  cross-lingual transferability of monolingual representations}.
\newblock In \emph{Proceedings of the 58th Annual Meeting of the Association
  for Computational Linguistics}, pages 4623--4637, Online. Association for
  Computational Linguistics.

\bibitem[{Bojanowski et~al.(2017)Bojanowski, Grave, Joulin, and
  Mikolov}]{bojanowski_enriching_2017}
Piotr Bojanowski, Edouard Grave, Armand Joulin, and Tomas Mikolov. 2017.
\newblock \href {https://doi.org/10.1162/tacl_a_00051} {Enriching {Word}
  {Vectors} with {Subword} {Information}}.
\newblock \emph{Transactions of the Association for Computational Linguistics},
  5:135--146.

\bibitem[{Chau et~al.(2020)Chau, Lin, and Smith}]{chau_parsing_2020}
Ethan~C. Chau, Lucy~H. Lin, and Noah~A. Smith. 2020.
\newblock \href {https://doi.org/10.18653/v1/2020.findings-emnlp.118} {Parsing
  with {Multilingual} {BERT}, a {Small} {Corpus}, and a {Small} {Treebank}}.
\newblock In \emph{Findings of the {Association} for {Computational}
  {Linguistics}: {EMNLP} 2020}, pages 1324--1334, Online. Association for
  Computational Linguistics.

\bibitem[{Chau and Smith(2021)}]{chau_specializing_2021}
Ethan~C. Chau and Noah~A. Smith. 2021.
\newblock \href {https://doi.org/10.18653/v1/2021.mrl-1.5} {Specializing
  multilingual language models: An empirical study}.
\newblock In \emph{Proceedings of the 1st Workshop on Multilingual
  Representation Learning}, pages 51--61, Punta Cana, Dominican Republic.
  Association for Computational Linguistics.

\bibitem[{Clark et~al.(2020)Clark, Luong, Le, and Manning}]{clark_electra_2020}
Kevin Clark, Minh-Thang Luong, Quoc~V. Le, and Christopher~D. Manning. 2020.
\newblock \href {https://doi.org/10.48550/arXiv.2003.10555} {{ELECTRA}:
  {Pre}-training {Text} {Encoders} as {Discriminators} {Rather} {Than}
  {Generators}}.

\bibitem[{Conneau et~al.(2020)Conneau, Khandelwal, Goyal, Chaudhary, Wenzek,
  Guzmán, Grave, Ott, Zettlemoyer, and Stoyanov}]{conneau_unsupervised_2020}
Alexis Conneau, Kartikay Khandelwal, Naman Goyal, Vishrav Chaudhary, Guillaume
  Wenzek, Francisco Guzmán, Edouard Grave, Myle Ott, Luke Zettlemoyer, and
  Veselin Stoyanov. 2020.
\newblock \href {https://doi.org/10.18653/v1/2020.acl-main.747} {Unsupervised
  {Cross}-lingual {Representation} {Learning} at {Scale}}.
\newblock In \emph{Proceedings of the 58th {Annual} {Meeting} of the
  {Association} for {Computational} {Linguistics}}, pages 8440--8451, Online.
  Association for Computational Linguistics.

\bibitem[{Devlin et~al.(2019)Devlin, Chang, Lee, and
  Toutanova}]{devlin_bert_2019}
Jacob Devlin, Ming-Wei Chang, Kenton Lee, and Kristina Toutanova. 2019.
\newblock \href {https://doi.org/10.18653/v1/N19-1423} {{BERT}: {Pre}-training
  of {Deep} {Bidirectional} {Transformers} for {Language} {Understanding}}.
\newblock In \emph{Proceedings of the 2019 {Conference} of the {North}
  {American} {Chapter} of the {Association} for {Computational} {Linguistics}:
  {Human} {Language} {Technologies}, {Volume} 1 ({Long} and {Short} {Papers})},
  pages 4171--4186, Minneapolis, Minnesota. Association for Computational
  Linguistics.

\bibitem[{Dozat and Manning(2017)}]{dozat_biaffine_2017}
Timothy Dozat and Christopher~D. Manning. 2017.
\newblock \href {https://openreview.net/forum?id=Hk95PK9le} {Deep biaffine
  attention for neural dependency parsing}.
\newblock In \emph{5th International Conference on Learning Representations,
  {ICLR} 2017, Toulon, France, April 24-26, 2017, Conference Track
  Proceedings}. OpenReview.net.

\bibitem[{Ebrahimi and Kann(2021)}]{ebrahimi_how_2021}
Abteen Ebrahimi and Katharina Kann. 2021.
\newblock \href {http://arxiv.org/abs/2106.02124} {How to {Adapt} {Your}
  {Pretrained} {Multilingual} {Model} to 1600 {Languages}}.
\newblock \emph{arXiv:2106.02124 [cs]}.

\bibitem[{Gardner et~al.(2018)Gardner, Grus, Neumann, Tafjord, Dasigi, Liu,
  Peters, Schmitz, and Zettlemoyer}]{gardner-etal-2018-allennlp}
Matt Gardner, Joel Grus, Mark Neumann, Oyvind Tafjord, Pradeep Dasigi,
  Nelson~F. Liu, Matthew Peters, Michael Schmitz, and Luke Zettlemoyer. 2018.
\newblock \href {https://doi.org/10.18653/v1/W18-2501} {{A}llen{NLP}: A deep
  semantic natural language processing platform}.
\newblock In \emph{Proceedings of Workshop for {NLP} Open Source Software
  ({NLP}-{OSS})}, pages 1--6, Melbourne, Australia. Association for
  Computational Linguistics.

\bibitem[{Goyal et~al.(2020)Goyal, Kunchukuttan, Kejriwal, Jain, and
  Bhagwat}]{goyal-etal-2020-contact}
Vikrant Goyal, Anoop Kunchukuttan, Rahul Kejriwal, Siddharth Jain, and Amit
  Bhagwat. 2020.
\newblock \href {https://aclanthology.org/2020.wmt-1.19} {Contact relatedness
  can help improve multilingual {NMT}: {M}icrosoft {STCI}-{MT} @ {WMT}20}.
\newblock In \emph{Proceedings of the Fifth Conference on Machine Translation},
  pages 202--206, Online. Association for Computational Linguistics.

\bibitem[{Houlsby et~al.(2019)Houlsby, Giurgiu, Jastrzebski, Morrone,
  de~Laroussilhe, Gesmundo, Attariyan, and
  Gelly}]{houlsby_parameter-efficient_2019}
Neil Houlsby, Andrei Giurgiu, Stanislaw Jastrzebski, Bruna Morrone, Quentin
  de~Laroussilhe, Andrea Gesmundo, Mona Attariyan, and Sylvain Gelly. 2019.
\newblock \href {http://arxiv.org/abs/1902.00751} {Parameter-{Efficient}
  {Transfer} {Learning} for {NLP}}.
\newblock \emph{arXiv:1902.00751 [cs, stat]}.

\bibitem[{Howard and Ruder(2018)}]{howard_universal_2018}
Jeremy Howard and Sebastian Ruder. 2018.
\newblock \href {https://doi.org/10.18653/v1/P18-1031} {Universal {Language}
  {Model} {Fine}-tuning for {Text} {Classification}}.
\newblock In \emph{Proceedings of the 56th {Annual} {Meeting} of the
  {Association} for {Computational} {Linguistics} ({Volume} 1: {Long}
  {Papers})}, pages 328--339, Melbourne, Australia. Association for
  Computational Linguistics.

\bibitem[{Joshi et~al.(2020)Joshi, Santy, Budhiraja, Bali, and
  Choudhury}]{joshi-etal-2020-state}
Pratik Joshi, Sebastin Santy, Amar Budhiraja, Kalika Bali, and Monojit
  Choudhury. 2020.
\newblock \href {https://doi.org/10.18653/v1/2020.acl-main.560} {The state and
  fate of linguistic diversity and inclusion in the {NLP} world}.
\newblock In \emph{Proceedings of the 58th Annual Meeting of the Association
  for Computational Linguistics}, pages 6282--6293, Online. Association for
  Computational Linguistics.

\bibitem[{Liu et~al.(2019)Liu, Ott, Goyal, Du, Joshi, Chen, Levy, Lewis,
  Zettlemoyer, and Stoyanov}]{liu_roberta_2019}
Yinhan Liu, Myle Ott, Naman Goyal, Jingfei Du, Mandar Joshi, Danqi Chen, Omer
  Levy, Mike Lewis, Luke Zettlemoyer, and Veselin Stoyanov. 2019.
\newblock \href {http://arxiv.org/abs/1907.11692} {{RoBERTa}: {A} {Robustly}
  {Optimized} {BERT} {Pretraining} {Approach}}.
\newblock \emph{arXiv:1907.11692 [cs]}.
\newblock ArXiv: 1907.11692.

\bibitem[{Martin et~al.(2020)Martin, Muller, Ortiz~Suárez, Dupont, Romary,
  de~la Clergerie, Seddah, and Sagot}]{martin_camembert_2020}
Louis Martin, Benjamin Muller, Pedro~Javier Ortiz~Suárez, Yoann Dupont,
  Laurent Romary, Éric de~la Clergerie, Djamé Seddah, and Benoît Sagot.
  2020.
\newblock \href {https://doi.org/10.18653/v1/2020.acl-main.645} {{CamemBERT}: a
  {Tasty} {French} {Language} {Model}}.
\newblock In \emph{Proceedings of the 58th {Annual} {Meeting} of the
  {Association} for {Computational} {Linguistics}}, pages 7203--7219, Online.
  Association for Computational Linguistics.

\bibitem[{McCann et~al.(2018)McCann, Bradbury, Xiong, and
  Socher}]{mccann_learned_2018}
Bryan McCann, James Bradbury, Caiming Xiong, and Richard Socher. 2018.
\newblock \href {http://arxiv.org/abs/1708.00107} {Learned in {Translation}:
  {Contextualized} {Word} {Vectors}}.
\newblock \emph{arXiv:1708.00107 [cs]}.
\newblock ArXiv: 1708.00107.

\bibitem[{Micallef et~al.(2022)Micallef, Gatt, and
  Tanti}]{micallef_pre-training_nodate}
Kurt Micallef, Albert Gatt, and Marc Tanti. 2022.
\newblock Pre-training {Data} {Quality} and {Quantity} for a {Low}-{Resource}
  {Language}: {New} {Corpus} and {BERT} {Models} for {Maltese}.
\newblock \emph{arXiv}, page~12.

\bibitem[{Micheli et~al.(2020)Micheli, d'Hoffschmidt, and
  Fleuret}]{micheli_importance_2020}
Vincent Micheli, Martin d'Hoffschmidt, and François Fleuret. 2020.
\newblock \href {https://doi.org/10.18653/v1/2020.emnlp-main.632} {On the
  importance of pre-training data volume for compact language models}.
\newblock In \emph{Proceedings of the 2020 {Conference} on {Empirical}
  {Methods} in {Natural} {Language} {Processing} ({EMNLP})}, pages 7853--7858,
  Online. Association for Computational Linguistics.

\bibitem[{Mikolov et~al.(2013{\natexlab{a}})Mikolov, Chen, Corrado, and
  Dean}]{mikolov_efficient_2013}
Tomas Mikolov, Kai Chen, Greg Corrado, and Jeffrey Dean. 2013{\natexlab{a}}.
\newblock \href {http://arxiv.org/abs/1301.3781} {Efficient {Estimation} of
  {Word} {Representations} in {Vector} {Space}}.
\newblock \emph{arXiv:1301.3781 [cs]}.
\newblock ArXiv: 1301.3781.

\bibitem[{Mikolov et~al.(2013{\natexlab{b}})Mikolov, Sutskever, Chen, Corrado,
  and Dean}]{mikolov_distributed_2013}
Tomas Mikolov, Ilya Sutskever, Kai Chen, Greg Corrado, and Jeffrey Dean.
  2013{\natexlab{b}}.
\newblock \href {http://arxiv.org/abs/1310.4546} {Distributed {Representations}
  of {Words} and {Phrases} and their {Compositionality}}.
\newblock \emph{arXiv:1310.4546 [cs, stat]}.
\newblock ArXiv: 1310.4546.

\bibitem[{Muller et~al.(2021)Muller, Anastasopoulos, Sagot, and
  Seddah}]{muller_when_2021}
Benjamin Muller, Antonios Anastasopoulos, Benoît Sagot, and Djamé Seddah.
  2021.
\newblock \href {https://doi.org/10.18653/v1/2021.naacl-main.38} {When {Being}
  {Unseen} from {mBERT} is just the {Beginning}: {Handling} {New} {Languages}
  {With} {Multilingual} {Language} {Models}}.
\newblock In \emph{Proceedings of the 2021 {Conference} of the {North}
  {American} {Chapter} of the {Association} for {Computational} {Linguistics}:
  {Human} {Language} {Technologies}}, pages 448--462, Online. Association for
  Computational Linguistics.

\bibitem[{Nguyen and Chiang(2017)}]{nguyen-chiang-2017-transfer}
Toan~Q. Nguyen and David Chiang. 2017.
\newblock \href {https://aclanthology.org/I17-2050} {Transfer learning across
  low-resource, related languages for neural machine translation}.
\newblock In \emph{Proceedings of the Eighth International Joint Conference on
  Natural Language Processing (Volume 2: Short Papers)}, pages 296--301,
  Taipei, Taiwan. Asian Federation of Natural Language Processing.

\bibitem[{Nivre et~al.(2016)Nivre, de~Marneffe, Ginter, Goldberg, Hajič,
  Manning, McDonald, Petrov, Pyysalo, Silveira, Tsarfaty, and
  Zeman}]{nivre_universal_2016}
Joakim Nivre, Marie-Catherine de~Marneffe, Filip Ginter, Yoav Goldberg, Jan
  Hajič, Christopher~D. Manning, Ryan McDonald, Slav Petrov, Sampo Pyysalo,
  Natalia Silveira, Reut Tsarfaty, and Daniel Zeman. 2016.
\newblock \href {https://aclanthology.org/L16-1262} {Universal {Dependencies}
  v1: {A} {Multilingual} {Treebank} {Collection}}.
\newblock In \emph{Proceedings of the {Tenth} {International} {Conference} on
  {Language} {Resources} and {Evaluation} ({LREC}'16)}, pages 1659--1666,
  Portorož, Slovenia. European Language Resources Association (ELRA).

\bibitem[{Ogueji et~al.(2021)Ogueji, Zhu, and Lin}]{ogueji_small_2021}
Kelechi Ogueji, Yuxin Zhu, and Jimmy Lin. 2021.
\newblock \href {https://aclanthology.org/2021.mrl-1.11} {Small {Data}? {No}
  {Problem}! {Exploring} the {Viability} of {Pretrained} {Multilingual}
  {Language} {Models} for {Low}-resourced {Languages}}.
\newblock In \emph{Proceedings of the 1st {Workshop} on {Multilingual}
  {Representation} {Learning}}, pages 116--126, Punta Cana, Dominican Republic.
  Association for Computational Linguistics.

\bibitem[{Ortiz~Suárez et~al.(2020)Ortiz~Suárez, Romary, and
  Sagot}]{ortiz_suarez_monolingual_2020}
Pedro~Javier Ortiz~Suárez, Laurent Romary, and Benoît Sagot. 2020.
\newblock \href {https://doi.org/10.18653/v1/2020.acl-main.156} {A
  {Monolingual} {Approach} to {Contextualized} {Word} {Embeddings} for
  {Mid}-{Resource} {Languages}}.
\newblock In \emph{Proceedings of the 58th {Annual} {Meeting} of the
  {Association} for {Computational} {Linguistics}}, pages 1703--1714, Online.
  Association for Computational Linguistics.

\bibitem[{Pan et~al.(2017)Pan, Zhang, May, Nothman, Knight, and
  Ji}]{pan-etal-2017-cross}
Xiaoman Pan, Boliang Zhang, Jonathan May, Joel Nothman, Kevin Knight, and Heng
  Ji. 2017.
\newblock \href {https://doi.org/10.18653/v1/P17-1178} {Cross-lingual name
  tagging and linking for 282 languages}.
\newblock In \emph{Proceedings of the 55th Annual Meeting of the Association
  for Computational Linguistics (Volume 1: Long Papers)}, pages 1946--1958,
  Vancouver, Canada. Association for Computational Linguistics.

\bibitem[{Paszke et~al.(2019)Paszke, Gross, Massa, Lerer, Bradbury, Chanan,
  Killeen, Lin, Gimelshein, Antiga, Desmaison, Kopf, Yang, DeVito, Raison,
  Tejani, Chilamkurthy, Steiner, Fang, Bai, and Chintala}]{NEURIPS2019_9015}
Adam Paszke, Sam Gross, Francisco Massa, Adam Lerer, James Bradbury, Gregory
  Chanan, Trevor Killeen, Zeming Lin, Natalia Gimelshein, Luca Antiga, Alban
  Desmaison, Andreas Kopf, Edward Yang, Zachary DeVito, Martin Raison, Alykhan
  Tejani, Sasank Chilamkurthy, Benoit Steiner, Lu~Fang, Junjie Bai, and Soumith
  Chintala. 2019.
\newblock \href
  {http://papers.neurips.cc/paper/9015-pytorch-an-imperative-style-high-performance-deep-learning-library.pdf}
  {Pytorch: An imperative style, high-performance deep learning library}.
\newblock In H.~Wallach, H.~Larochelle, A.~Beygelzimer, F.~d\textquotesingle
  Alch\'{e}-Buc, E.~Fox, and R.~Garnett, editors, \emph{Advances in Neural
  Information Processing Systems 32}, pages 8024--8035. Curran Associates, Inc.

\bibitem[{Pennington et~al.(2014)Pennington, Socher, and
  Manning}]{pennington_glove_2014}
Jeffrey Pennington, Richard Socher, and Christopher Manning. 2014.
\newblock \href {https://doi.org/10.3115/v1/D14-1162} {{GloVe}: {Global}
  {Vectors} for {Word} {Representation}}.
\newblock In \emph{Proceedings of the 2014 {Conference} on {Empirical}
  {Methods} in {Natural} {Language} {Processing} ({EMNLP})}, pages 1532--1543,
  Doha, Qatar. Association for Computational Linguistics.

\bibitem[{Peters et~al.(2018)Peters, Neumann, Iyyer, Gardner, Clark, Lee, and
  Zettlemoyer}]{peters_deep_2018}
Matthew~E. Peters, Mark Neumann, Mohit Iyyer, Matt Gardner, Christopher Clark,
  Kenton Lee, and Luke Zettlemoyer. 2018.
\newblock \href {https://doi.org/10.18653/v1/N18-1202} {Deep {Contextualized}
  {Word} {Representations}}.
\newblock In \emph{Proceedings of the 2018 {Conference} of the {North}
  {American} {Chapter} of the {Association} for {Computational} {Linguistics}:
  {Human} {Language} {Technologies}, {Volume} 1 ({Long} {Papers})}, pages
  2227--2237, New Orleans, Louisiana. Association for Computational
  Linguistics.

\bibitem[{Pfeiffer et~al.(2020{\natexlab{a}})Pfeiffer, Rücklé, Poth, Kamath,
  Vulić, Ruder, Cho, and Gurevych}]{pfeiffer_adapterhub_2020}
Jonas Pfeiffer, Andreas Rücklé, Clifton Poth, Aishwarya Kamath, Ivan Vulić,
  Sebastian Ruder, Kyunghyun Cho, and Iryna Gurevych. 2020{\natexlab{a}}.
\newblock \href {https://doi.org/10.18653/v1/2020.emnlp-demos.7} {{AdapterHub}:
  {A} {Framework} for {Adapting} {Transformers}}.
\newblock In \emph{Proceedings of the 2020 {Conference} on {Empirical}
  {Methods} in {Natural} {Language} {Processing}: {System} {Demonstrations}},
  pages 46--54, Online. Association for Computational Linguistics.

\bibitem[{Pfeiffer et~al.(2020{\natexlab{b}})Pfeiffer, Vulić, Gurevych, and
  Ruder}]{pfeiffer_mad-x_2020}
Jonas Pfeiffer, Ivan Vulić, Iryna Gurevych, and Sebastian Ruder.
  2020{\natexlab{b}}.
\newblock \href {https://doi.org/10.18653/v1/2020.emnlp-main.617} {{MAD}-{X}:
  {An} {Adapter}-{Based} {Framework} for {Multi}-{Task} {Cross}-{Lingual}
  {Transfer}}.
\newblock In \emph{Proceedings of the 2020 {Conference} on {Empirical}
  {Methods} in {Natural} {Language} {Processing} ({EMNLP})}, pages 7654--7673,
  Online. Association for Computational Linguistics.

\bibitem[{Pires et~al.(2019)Pires, Schlinger, and Garrette}]{pires_how_2019}
Telmo Pires, Eva Schlinger, and Dan Garrette. 2019.
\newblock \href {https://doi.org/10.18653/v1/P19-1493} {How {Multilingual} is
  {Multilingual} {BERT}?}
\newblock In \emph{Proceedings of the 57th {Annual} {Meeting} of the
  {Association} for {Computational} {Linguistics}}, pages 4996--5001, Florence,
  Italy. Association for Computational Linguistics.

\bibitem[{Raffel et~al.(2020)Raffel, Shazeer, Roberts, Lee, Narang, Matena,
  Zhou, Li, and Liu}]{raffel_exploring_2020}
Colin Raffel, Noam Shazeer, Adam Roberts, Katherine Lee, Sharan Narang, Michael
  Matena, Yanqi Zhou, Wei Li, and Peter~J. Liu. 2020.
\newblock \href {http://arxiv.org/abs/1910.10683} {Exploring the {Limits} of
  {Transfer} {Learning} with a {Unified} {Text}-to-{Text} {Transformer}}.

\bibitem[{Rehurek and Sojka(2011)}]{rehurek2011gensim}
Radim Rehurek and Petr Sojka. 2011.
\newblock Gensim--python framework for vector space modelling.
\newblock \emph{NLP Centre, Faculty of Informatics, Masaryk University, Brno,
  Czech Republic}, 3(2).

\bibitem[{Riabi et~al.(2021)Riabi, Sagot, and Seddah}]{riabi_can_2021}
Arij Riabi, Benoît Sagot, and Djamé Seddah. 2021.
\newblock \href {http://arxiv.org/abs/2110.13658} {Can {Character}-based
  {Language} {Models} {Improve} {Downstream} {Task} {Performance} in
  {Low}-{Resource} and {Noisy} {Language} {Scenarios}?}

\bibitem[{Rogers et~al.(2020)Rogers, Kovaleva, and
  Rumshisky}]{rogers_primer_2020-3}
Anna Rogers, Olga Kovaleva, and Anna Rumshisky. 2020.
\newblock \href {http://arxiv.org/abs/2002.12327} {A {Primer} in {BERTology}:
  {What} we know about how {BERT} works}.
\newblock \emph{arXiv:2002.12327 [cs]}.
\newblock ArXiv: 2002.12327 version: 3.

\bibitem[{Schroeder and Zeldes(2016)}]{schroeder_raiders_2016}
Caroline~T. Schroeder and Amir Zeldes. 2016.
\newblock \href
  {http://www.digitalhumanities.org/dhq/vol/10/2/000247/000247.html} {Raiders
  of the {Lost} {Corpus}}.
\newblock \emph{Digital Humanities Quarterly}, 010(2).

\bibitem[{Turc et~al.(2019)Turc, Chang, Lee, and
  Toutanova}]{turc_well-read_2019}
Iulia Turc, Ming-Wei Chang, Kenton Lee, and Kristina Toutanova. 2019.
\newblock \href {https://doi.org/10.48550/arXiv.1908.08962} {Well-{Read}
  {Students} {Learn} {Better}: {On} the {Importance} of {Pre}-training
  {Compact} {Models}}.
\newblock ArXiv:1908.08962 [cs].

\bibitem[{Vaswani et~al.(2017)Vaswani, Shazeer, Parmar, Uszkoreit, Jones,
  Gomez, Kaiser, and Polosukhin}]{vaswani_attention_2017}
Ashish Vaswani, Noam Shazeer, Niki Parmar, Jakob Uszkoreit, Llion Jones,
  Aidan~N. Gomez, Łukasz Kaiser, and Illia Polosukhin. 2017.
\newblock \href
  {https://papers.nips.cc/paper/2017/hash/3f5ee243547dee91fbd053c1c4a845aa-Abstract.html}
  {Attention is {All} you {Need}}.
\newblock \emph{Advances in Neural Information Processing Systems}, 30.

\bibitem[{Vatri and McGillivray(2018)}]{vatri_diorisis_2018}
A.~Vatri and B.~McGillivray. 2018.
\newblock \href {https://doi.org/https://doi.org/10.1163/24523666-01000013}
  {The {Diorisis} {Ancient} {Greek} {Corpus}: {Linguistics} and {Literature}}.
\newblock \emph{Research Data Journal for the Humanities and Social Sciences},
  3(1):55 -- 65.
\newblock Place: Leiden, The Netherlands Publisher: Brill.

\bibitem[{Wang et~al.(2020)Wang, K, Mayhew, and Roth}]{wang_extending_2020}
Zihan Wang, Karthikeyan K, Stephen Mayhew, and Dan Roth. 2020.
\newblock \href {http://arxiv.org/abs/2004.13640} {Extending {Multilingual}
  {BERT} to {Low}-{Resource} {Languages}}.
\newblock \emph{arXiv:2004.13640 [cs]}.
\newblock ArXiv: 2004.13640.

\bibitem[{Warstadt et~al.(2020)Warstadt, Zhang, Li, Liu, and
  Bowman}]{warstadt_learning_2020}
Alex Warstadt, Yian Zhang, Xiaocheng Li, Haokun Liu, and Samuel~R. Bowman.
  2020.
\newblock \href {https://doi.org/10.18653/v1/2020.emnlp-main.16} {Learning
  {Which} {Features} {Matter}: {RoBERTa} {Acquires} a {Preference} for
  {Linguistic} {Generalizations} ({Eventually})}.
\newblock In \emph{Proceedings of the 2020 {Conference} on {Empirical}
  {Methods} in {Natural} {Language} {Processing} ({EMNLP})}, pages 217--235,
  Online. Association for Computational Linguistics.

\bibitem[{Wolf et~al.(2020)Wolf, Debut, Sanh, Chaumond, Delangue, Moi, Cistac,
  Rault, Louf, Funtowicz, and Brew}]{wolf_huggingfaces_2020}
Thomas Wolf, Lysandre Debut, Victor Sanh, Julien Chaumond, Clement Delangue,
  Anthony Moi, Pierric Cistac, Tim Rault, Rémi Louf, Morgan Funtowicz, and
  Jamie Brew. 2020.
\newblock \href {http://arxiv.org/abs/1910.03771} {{HuggingFace}'s
  {Transformers}: {State}-of-the-art {Natural} {Language} {Processing}}.
\newblock \emph{arXiv:1910.03771 [cs]}.
\newblock ArXiv: 1910.03771.

\bibitem[{Wu and Dredze(2020)}]{wu_are_2020}
Shijie Wu and Mark Dredze. 2020.
\newblock \href {https://doi.org/10.18653/v1/2020.repl4nlp-1.16} {Are {All}
  {Languages} {Created} {Equal} in {Multilingual} {BERT}?}
\newblock In \emph{Proceedings of the 5th {Workshop} on {Representation}
  {Learning} for {NLP}}, pages 120--130, Online. Association for Computational
  Linguistics.

\end{thebibliography}
\bibliographystyle{acl_natbib}

\clearpage
\appendix
\section{Acknowledgments}
We thank Nathan Schneider, Shabnam Tafreshi, and MASC-SLL 2022 attendees for very helpful comments on this work.

\section{Datasets}
\label{sec:datasets}
\begin{table}[ht]
    \centering
    \begin{tabular}{l|r}
        Treebank & Tokens  \\\hline
        UD\_Coptic-Scriptorium v2.9    &  48,632 \\
        UD\_Ancient\_Greek-PROEIL v2.9 & 213,999 \\
        UD\_Indonesian-GSD v2.10       & 122,021 \\
        UD\_Maltese-MUDT v2.9          &  44,162 \\
        UD\_Uyghur-UDT v2.9            &  40,236 \\
        UD\_Wolof-WDT v2.9             &  44,258 \\
        UD\_Tamil-TTB v2.10            &   9,581 \\
    \end{tabular}
    \caption{Token count statistics for UD treebanks used in this work. Note that for this count, we count the constituent tokens of multiword tokens instead of counting a multiword token as a single token.}
    \label{tab:ud_stats}
\end{table}

\paragraph{Unlabeled}
For Coptic, we use v4.2.0 of the Coptic SCRIPTORIUM corpora \citep{schroeder_raiders_2016}, obtained from \url{https://github.com/copticscriptorium/corpora}.
For Ancient Greek, we use the initial release of the Diorisis corpus \citep{vatri_diorisis_2018}, obtained from \url{https://figshare.com/articles/dataset/The_Diorisis_Ancient_Greek_Corpus/6187256}.
Both corpora are preprocessed (tokenized, etc.) using language-specific tools to a quality higher than would have been obtained with a generic preprocessing pipeline.
In Coptic's case, the data is further checked and with parts gold annotated by humans. 

All other corpora are derived from Wikipedia. 
For Maltese, Uyghur, and Wolof, we use all available namespace 0 articles\footnote{Wikipedia articles belonging to namespace 0 are main content articles instead of e.g.~user pages or template pages.} as of February 2022, and for Indonesian and Tamil, we take a random sampling of namespace 0 articles as of June 2022, up to around 1.5M tokens. 

All data is derived from Wikipedia's public dump files.
While it is popular in NLP to use the text in the dump files directly, this is suboptimal, as the dump files' text contains markup, which makes the text noisy and means that document structural information cannot be used in the tokenization and sentence splitting process. 
We therefore take the additional step of rendering the dump into HTML using \Thresher{}, which can then be used to obtain useful information about guaranteed sentence splits, e.g.~between HTML elements like \texttt{<p>}.
We perform rule-based sentence splitting and tokenization on this HTML to obtain our final tokenized texts.

For all 7 languages, we reserve around 10\% of documents for validation and use the rest for training.
A test split is unnecessary because our models are not being evaluated on unlabeled data.

\paragraph{UD Treebanks}
A summary of the treebanks we use and their versions is given in \Cref{tab:ud_stats}.
We use the standard train/dev/test splits for all treebanks.

\paragraph{WikiAnn Datasets}
New train/dev/test splits were created in an 8:1:1 ratio for the WikiAnn dataset, which only divides sentences by language.
It was not possible to split at the document level because no document metadata is available in the WikiAnn dataset.
Tags are converted from the native IOB1 scheme into the BIOUL scheme.
Some manual edits, logged in our version control history, were made to sentence boundaries in order to keep wordpiece sequence lengths below 512.

\section{Conditions}
\label{sec:conditions}
All experiments for both pretraining and evaluation were performed on NVIDIA Tesla T4 GPUs with 16GB GDDR6 SDRAM. 

\paragraph{Word2vec}
We use the Gensim \citep{rehurek2011gensim} implementation of the Word2vec skip-gram with negative sampling algorithm for pretrained static word embeddings.
The embeddings are trained just on the train split of the unlabeled corpus for each language.
The vectors are 100-dimensional, window size is 5, and negative sampling factor is 5. 

\paragraph{mBERT-VA}
We implement the Vocabulary Augmentation method exactly as prescribed by \citet{chau_parsing_2020} by training a new wordpiece tokenizer on the train split of the unlabeled data with a vocabulary size of 5,000, yielding a new monolingual vocabulary.
The monolingual vocabulary is ranked by frequency of wordpieces, and the 99 unused tokens in mBERT's vocabulary indexed between 1 and 99 are replaced by tokens from the monolingual vocabulary which are not already present in mBERT's vocabulary.
Since only preexisting token indices are used, it is not necessary to modify the model's pretrained weights.

To train the weights of the previously unused token indices, adaptive pretraining with MLM is performed, again following \citet{chau_parsing_2020}.
Due to memory constraints on our GPUs, we are forced to set the batch size to 2.
We pretrain for 20 epochs with 16,000 batches per epoch.
The PyTorch AdamW optimizer is used with $\beta_1=0.9, \beta_2=0.999$, learning rate at 1e-4, and weight decay at 0.05.
The model which achieved lowest validation set perplexity is chosen.

\paragraph{MicroBERT}
Our tokenizer for MicroBERT is a WordPiece tokenizer.
We scale vocabulary size from a minimum of 8,000 wordpieces up to 14,000 wordpieces, where the number of unique whitespace tokens for a given language determines how large the vocabulary will be.
All models are uncased and perform Unicode NFD normalization as a preprocessing step during tokenization.

Since some tasks require wordpieces while others require tokens (e.g. PoS tagging), our encoder produces both wordpiece sequences and token sequences.
The token sequence is constructed by keeping track of which wordpieces correspond to which original input tokens, and average pooling wordpieces for each token so that the sequence length reflects the number of original input tokens.

During data loading, sequences longer than 500 wordpieces are split into chunks of no more than 500 wordpieces each.
Sequences this long only occur in the unlabeled datasets, so this does not pose a problem for producing valid losses on PoS tagging or parsing.

We train with a batch size of 32 for 200 epochs with 8,000 batches per epoch.
We again use the AdamW optimizer with a learning rate of 3e-3, $\beta_1=0.9$, $\beta_2=0.999$, and weight decay at 0.05. 
We allow early stopping if the validation metric, MLM perplexity, shows no improvement for 40 epochs.
The model with the best validation MLM perplexity is selected.

While it is traditionally popular to train BERTs with triangular learning rates \citep{howard_universal_2018}, we chose not to use them for training our MicroBERTs.
The reason is that, as noted by \citet{raffel_exploring_2020}, it is necessary to know in advance approximately how many training steps are necessary to train a model, but since our MicroBERT architecture is much smaller, it is not obvious how many steps would be required to train it, making its use difficult.
We do not expect this to lead to much worse performance compared to a properly configured triangular learning rate, as \citet{raffel_exploring_2020} also note that the triangular schedule often leads to only marginal gains compared to other schedules.
Instead, we use PyTorch's ReduceLROnPlateau scheduler, which reduces learning rate when a certain number of validation steps have shown no improvement in MLM perplexity.
We configure the scheduler so that if no improvement occurs for 2 epochs, the learning rate is halved, down to a minimum learning rate of 5e-5.
Our results have shown that this training regimen can achieve good results, but we expect there is room for improvement and leave the task of refining it to future work.

\section{Evaluation}
\label{sec:evaluation}

\paragraph{Parsing} 
We use the AllenNLP implementation of a biaffine attention parser \citep{dozat_biaffine_2017}.
In line with previous work, we set the dimensionality of the arc and tag representations to 100, and dropout and input dropout are set to 0.3.
An encoder stack of 3 bidirectional LSTMs is used, with a recurrent dropout of 0.3, hidden size of 400, and highway connections. 
A scalar mix of representations from each layer of the BERT model is learned \citep{peters_deep_2018} to allow the model to fully exploit information present in earlier layers. 
Gold tokenization is used, and no supplementary representations (such as static word embeddings or feature or PoS embeddings) are provided. 

We train for 300 epochs with a batch size of 16 and patience of 50 with LAS as our validation metric. 
To account for the very large size of some treebanks (e.g.~Greek), we train for 200 batches per epoch.
The AdamW optimizer is used with $\beta_1=0.9$, $\beta_2=0.999$, learning rate at 1e-3, and gradient clipping at 5.0. 
A reduced learning rate of 5e-5 is used for all parameters in the TLM. 

\paragraph{NER}
We use AllenNLP's linear chain CRF tagger with BIOUL encoding. 
As with parsing, a scalar mix of representations from each layer of the BERT model is learned \citep{peters_deep_2018} to allow the model to fully exploit information present in earlier layers. 
An encoder stack of 2 bidirectional LSTMs is used, with a dropout of 0.5 and hidden size of 200.
The model's dropout is set to 0.5.
Gold tokenization is used, and no supplementary representations (such as static word embeddings or feature or PoS embeddings) are provided.

We train for 300 epochs with a batch size of 16 and patience of 50 with span-based F1 as our validation metric. 
The AdamW optimizer is used with $\beta_1=0.9$, $\beta_2=0.999$, learning rate at 1e-3, and gradient clipping at 5.0. 
A reduced learning rate of 1e-5 is used for all parameters in the TLM.

\end{document}